\documentclass{article}
\usepackage{float}

\PassOptionsToPackage{numbers}{natbib}
\usepackage[preprint]{nips_2018}

\usepackage{graphicx}
\usepackage{wrapfig}

\usepackage[utf8]{inputenc} % allow utf-8 input
\usepackage[T1]{fontenc}    % use 8-bit T1 fonts
\usepackage{hyperref}       % hyperlinks
\usepackage{url}            % simple URL typesetting
\usepackage{booktabs}       % professional-quality tables
\usepackage{amsfonts}       % blackboard math symbols
\usepackage{nicefrac}       % compact symbols for 1/2, etc.
\usepackage{microtype}      % microtypography
\usepackage{amsmath}
\usepackage{soul}

\usepackage{xargs}   
\usepackage[colorinlistoftodos,prependcaption,textsize=tiny]{todonotes}
\newcommandx{\leon}[2][1=]{\todo[linecolor=red,backgroundcolor=red!25,bordercolor=red,#1]{#2}}

\newcommandx{\hugo}[2][1=]{\todo[linecolor=green,backgroundcolor=white!25,bordercolor=red,#1]{#2}}

\newcommandx{\andrew}[2][1=]{\todo[linecolor=blue,backgroundcolor=white!25,bordercolor=red,#1]{#2}}

\title{Multitaper Spectral Estimation HDP-HMMs for EEG Sleep Inference}

\author{
  Leon Chlon\thanks{These authors contributed equally to this work.}\\
  MIT Department of BCS\\
  Cambridge, MA 02139\\
  \texttt{lchlon@mit.edu}\And
  Andrew H. Song$^{*}$\\
  MIT Department of EECS\\
  Cambridge, MA 02139\\
  \texttt{andrew90@mit.edu}\And
  Sandya Subramanian\\
  Harvard-MIT HST\\
  Cambridge, MA 02139\\
  \texttt{sandya@mit.edu}\And
  Hugo Soulat\\
  MIT Department of BCS\\
  Cambridge, MA 02139\\
  \texttt{hsoulat@mit.edu}\And
  John Tauber\\
  MIT Department of BCS\\
  Cambridge, MA 02139\\
  \texttt{jtauber@mit.edu}\And
  Demba Ba\\
  SEAS Department of Harvard\\
  Cambridge, MA 02138\\
  \texttt{demba@seas.harvard.edu}\And
   Michael Prerau\\
MGH Department of Anesthesia, Critical Care and Pain Medicine\\
55 Fruit st, GRJ 4, Boston, MA 02114\\
 \texttt{prerau@nmr.mgh.harvard.edu}
}

\begin{document}

\maketitle

\begin{abstract}
Electroencephalographic (EEG) monitoring of neural activity is widely used for sleep disorder diagnostics and research. The standard of care is to manually classify 30-second epochs of EEG time-domain traces into 5 discrete sleep stages. Unfortunately, this scoring process is subjective and time-consuming, and the defined stages do not capture the heterogeneous landscape of healthy and clinical neural dynamics. This motivates the search for a data-driven and principled way to identify the number and composition of salient, reoccurring brain states present during sleep. To this end, we propose a Hierarchical Dirichlet Process Hidden Markov Model (HDP-HMM), combined with wide-sense stationary (WSS) time series spectral estimation to construct a generative model for personalized subject sleep states. In addition, we employ multitaper spectral estimation to further reduce the large variance of the spectral estimates inherent to finite-length EEG measurements. By applying our method to both simulated and human sleep data, we arrive at three main results: 1) a Bayesian nonparametric automated algorithm that recovers general temporal dynamics of sleep, 2) identification of subject-specific "microstates" within canonical sleep stages, and 3) discovery of stage-dependent sub-oscillations with shared spectral signatures across subjects. 
\end{abstract}

%\listoftodos[Notes]

\section{Introduction}

During sleep, the brain displays highly heterogeneous cortical oscillatory dynamics consisting of a complex interplay of numerous networks related to arousal and loss of consciousness \cite{Grigg-Damberger2012TheLater.,Loomis1937CerebralPotentials.,Brown2012ControlWakefulness}. The current clinical standard uses 30-second epochs of electroencephalogram (EEG) time-domain traces to categorize brain state during sleep into 5 discrete sleep stages: wake, rapid eye movement (REM) sleep, and non-REM sleep, which consists of 3 sub-stages, notated NREM1 through NREM3 \cite{Kales1968ASubjects}. The progression of these sleep stages through the night, called a hypnogram, is used to characterize sleep architecture. However, as these stages are determined subjectively through visual inspection by trained sleep technicians, this process is both time-consuming and inherently suffers from $\sim$ $20\%$ inter-scorer variability \cite{Grigg-Damberger2012TheLater.,Prerau2017SleepAnalysis}, which is greatly exacerbated in the case of pathological sleep \cite{Silber2007TheAdults.}. 

The chief explanation for this variability is that sleep, in every dimension thus far studied, is a continuous and dynamic process \cite{Ogilvie2001TheAsleep}, which is cleaved into a low-resolution, discrete, state-space by clinical staging. Furthermore, the current clinical standard is based on features easily observed in the time-domain by eye, and limited to a crude time and state resolution that is designed to reduce variability and scoring time in subjective visual categorization, rather than to maximize information content. Recent studies have highlighted the information-rich nature of neural oscillatory time-frequency dynamics within the sleep EEG that can observed through the multitaper spectrogram \cite{Prerau2017SleepAnalysis}. While there is indeed a continuum of changes in the sleep EEG, the identification of relevant and recurring combinations of oscillatory activity is vital for enhancing our understanding of the interaction of neural mechanisms underlying sleep, as well as pathophysiological deviations from the norm. It is therefore useful to devise a method for parcellating sleep, which is data rather than semantically driven, incorporates time-frequency dynamics, and can determine the state resolution without limitations imposed by the need for human scorers or a pre-assumed number of states. 

Unsupervised machine learning methods are ideal candidates for this class of problem. However, existing parametric approaches including Deep Belief Network - Hidden Markov Models (DBN-HMMs) \cite{Langkvist2012SleepLearning}, k-Nearest Neighbor (k-NN)-based methods \cite{Gunes2010EfficientWeighting}, Conditional Random Fields (CRFs) \cite{Luo2007Subject-adaptiveField.} and others \cite{Gath1994UnsupervisedPatterns,Sunagawa2013FASTER:Mice} constrain the total number of states. %A more desirable method would ascertain the number of states and their defining characteristics in a data-driven manner, and allow the flexibility for these states to differ between individuals. This would grant sleep researchers a powerful tool for exploring sleep dynamics across a heterogeneous population and the flexibility to identify and examine potential microstates of sleep that may have physiological importance. [TALK ABOUT THE NEED FOR A BAYESIAN FRAMEWORK HERE] %Despite these encouraging %removed : impressive
%results, these methods do not comprehensively address the outlined shortcomings in model inflexibility and information loss. The most concerning issues arise with respect to estimation of spectral characteristics from raw EEG, which gives rise to biased and noisy estimates of the true power spectrum \cite{Babadi2014AAnalysis}. 1) Simply averaging the power across the canonical frequency bands minimizes variability of the estimate, but it leads to a poor, low-resolution characterization of the power spectrum, with phase information loss. 2) The nonparametric HMM proposed by \cite{UlrichAnalysisTime-Series} neglects to address the issue of inaccurate spectral estimation, mixing uncertainty into their estimates of rodent sleep stages, and subsequently, any claims of the underlying neurophysiology
To address this shortcoming, we propose a Bayesian nonparametric framework that integrates 1) the Hierarchical Dirichlet Process Hidden Markov Model (HDP-HMM) \cite{Beal2002TheModel} for the underlying state dynamics and, 2) spectral representations and asymptotic properties of the wide-sense stationary (WSS) time series for the emission distribution of the generative model. %Specifically, we implement a Hierarchical Dirichlet Process Hidden Markov Model (HDP-HMM) over 15 second non-overlapping bins across each subject's computed spectrogram to infer a flexible set of sleep state assignments and transition probabilities. 
The multitaper spectral estimation framework \cite{Babadi2014AAnalysis} is incorporated to further optimize the bias and variance of the estimates of spectral characteristics for each state, compared to the estimates based on simple Fourier coefficients of the observations. Finally, we use the symmetrized form of the Kullback–Leibler (KL) divergence \cite{Kullback1987TheDistance} to cluster the inferred states across our patient cohort, with interpretation provided by subject-matter experts. To our knowledge, ours is the first HDP-HMM to applied to human sleep EEG data, along with the multitaper framework. Our work %method 
is readily extensible beyond sleep inference to other domains of neuroscience, such as nonparametric state modeling of the neurophysiology underlying epilepsy or Parkinson's disease.

\section{Model}
If the discrete time series, $\{y_t\}_t$, can be assumed to be realizations of a wide-sense stationary (WSS) stochastic process, we can apply several useful properties for parameter estimation and inference in time series analysis. WSS, or second-order stationarity, states that the mean and the autocovariance of the stochastic process is invariant with respect to time $t$. In practice, since the long time series ($\sim$ several hours of data for our study) may demonstrate nonstationarity, we segment the data into multiple non-overlapping windows, within which we assume WSS.

Let $N$ be the number of sample points for the entire time series and $J$ the number of samples in each window. Then,  T= $\lfloor{\frac{N}{J} \rfloor} $ denotes the number of windows for the data. With $t=1,\cdots,T$ as the index for a window, we define $y_t=\{y_{(t-1)J},\cdots,y_{tJ-1}\}$ as the samples in the window $t$.

\subsection{Spectral Representation for Wide-Sense Stationary Time Series}\label{section:WSS}
To analyze the discretely sampled time series with sampling rate $F_s$ in the frequency domain, we use the Discrete Fourier Transform (DFT) \cite{Oppenheim2010Discrete-timeProcessing}. Denoting $y_t^{(F)}\in \mathbb{C}^J$ as the DFT coefficients of $y_t\in \mathbb{R}^J$,
\begin{equation}\label{eq:DFT}
y_t^{(F)}(w_j) = \frac{1}{J}\sum_{l=1}^{J} y_{(t-1)J+l}W_J^{(j-1)(l-1)}
\end{equation}
where $W_J=\exp\big(-i\frac{2\pi}{J}\big)$ and $w_j=\frac{j-1}{J}$ for $j=1,\cdots,\frac{J}{2}$ is the normalized frequency. The conversion between frequency in Hz, $f_j$, and $w_j$ is given as $f_j=F_sw_j$. We will use $w_j$ to refer the frequency index throughout this work. Also for notational convenience, we will use $d_{j,t}^{Re}=Re\big\{y_t^{(F)}(w_j)\big\}$ and $d_{j,t}^{Im} = Im\big\{y_t^{(F)}(w_j)\big\}$. We now introduce the following lemma \cite{Brillinger2001TimeTheory}:

\textbf{Proposition 1.} If $y_t$ is a WSS time series, the DFT coefficients, $d_{j,t}^{Re}$ and $d_{j,t}^{Im}$, are distributed as asymptotically independent normal, as $J\rightarrow \infty$,
\begin{equation}\label{eq:an}
d_{j,t}^{Re}\sim \mathcal{N}\Big(0,\frac{f(w_j)}{2}\Big)\,\,\,\text{and}\,\,\,d_{j,t}^{Im}\sim \mathcal{N}\Big(0,\frac{f(w_j)}{2}\Big)
\end{equation}
where $f(w_j)$ is the true underlying PSD of the stochastic process at $w_j$ \cite{Brillinger2001TimeTheory} (for further explanation, see Supplementary Material Section 1). The independence property is the guiding principle for our generative model in section \ref{section:generative_model} and clustering analysis in section \ref{section:clustering}. 

\subsection{Multitaper Spectral Estimation}
The spectral estimation problem, or the problem of estimating the true PSD, $f(w_j)$, from observations (DFT coefficients), has received great attention. The simplest estimate, albeit with high bias and variance, is $\widehat{f}(w_j)=\lVert y_t^{(F)}(w_j)\rVert^2$, the \textit{periodogram}. Numerous methods have attempted to lower both bias and variance of $\widehat{f}(w_j)$ \cite{Percival1993SpectralTechniques}. Multitaper spectral estimation \cite{Thomson1982SpectrumAnalysis} optimizes the reduction in bias and variance of the PSD estimate by applying a set of specific tapers, or windowing functions, to the observed time series to obtain $M$ pseudo-observations, $\{y_t^{(m,F)}(w_j)\}_{m=1}^M$, 
\begin{equation*}
y_t^{(m,F)}(w_j)= \frac{1}{J}\sum_{l=1}^{J} h_l^{(m)}y_{(t-1)J+l}W_J^{(j-1)(l-1)}
\end{equation*}
where $h_l^{(m)}\in \mathbb{R}^J$ is the $m^{th}$ taper, $M$ being the number of tapers. The discrete prolate spheroidal sequences, which are mutually orthogonal with optimal energy concentration properties, are used as tapers. With these tapers, the periodogram estimates formed from the pseudo-observations, $\{\lVert y_t^{(m,F)}(w_j) \rVert^2\}_{m=1}^M$, are approximately uncorrelated \cite{Babadi2014AAnalysis}. Therefore, we can write the final estimator, $\widehat{f}_{MT}(w_j)$, as,
\begin{equation}
\widehat{f}_{MT}(w_j) = \frac{1}{M}\sum_{m=1}^M \lVert y_t^{(m,F)}(w_j) \rVert^2
\end{equation}
Figure 1 shows an example of the multitapered spectrogram and PSD for a segment of sleep EEG.

\begin{figure}\label{fig:generative}
  \centering
  \includegraphics[width=\linewidth, trim= 0cm 0cm 0cm 0cm, clip]{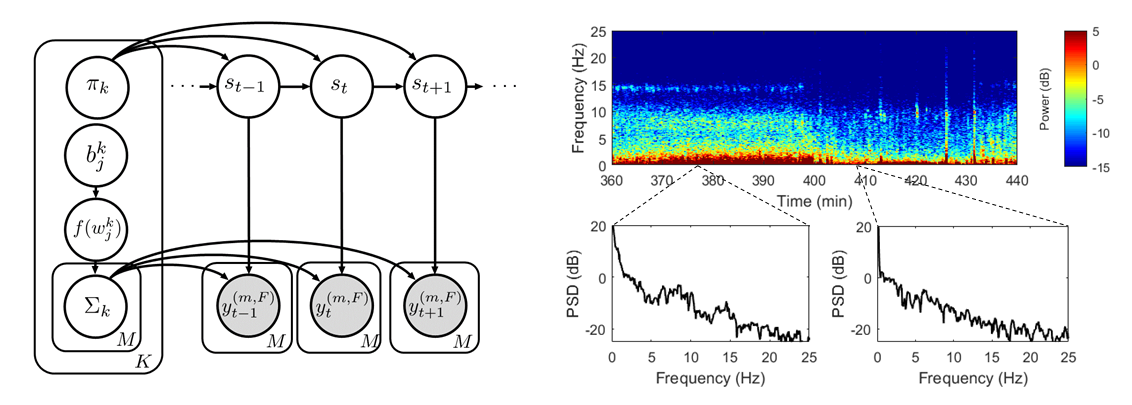}
  \caption{\label{fig:gm_spectrogram} The generative model for the HDP-HMM and example spectrogram.  Left : Graphical Model of HDP-HMM. Top-Right : Multi-tapered spectrogram of 80 minutes of sleep EEG data. Bottom right :  extracted PSD of 15 second windows. }
\end{figure}

\subsection{Generative model}\label{section:generative_model}
Our generative model, as depicted in Figure \ref{fig:gm_spectrogram}, follows an HDP-HMM framework \cite{Beal2002TheModel}. The observations, or DFT coefficients for each time window across frequency bands of interest, are generated by the \textit{spectral representation emission model}. We denote the spectral representation at each time window $t$ as $y_t^{(m,F)}(w_j)$, where $t=1,\cdots,T$ is the time window, $m=1,\cdots,M$ the tapers, and $w_j$ for $j=1,\cdots,\frac{J}{2}$ is the frequency index.

\subsubsection{Spectral Representation Emission Model}
We use the normal distribution in eq.(\ref{eq:an}) as the likelihood, $p\Big(y_t^{(F)} \Big\vert \theta^{(s_t)}\Big)$, where $\theta^{(s_t)}=\{f(w_j^{(s_t)}) \}_{j}$,
\begin{equation}
\begin{split}
&\big[d_{1,t}^{Re}, d_{1,t}^{Im},\cdots,d_{\frac{J}{2},t}^{Re}, d_{\frac{J}{2},t}^{Im}\big]\sim p\bigg(y_t^{(F)}\Big\vert \theta^{(s_t)}\bigg)=\mathcal{N}\big(\bar{\text{\bf{0}}},\Sigma_{s_t}\big)=\prod_{j=1}^{\frac{J}{2}}\Big\{\mathcal{N}\Big(0,\frac{f(w_j^{(s_t)})}{2}\Big)\Big\}^2\\
\end{split}
\end{equation}
where $\Sigma_{s_t}=\frac{1}{2}diag(f(w_1^{(s_t)}),f(w_1^{(s_t)}),\cdots,f(w_{\frac{J}{2}}^{(s_t)}),f(w_{\frac{J}{2}}^{(s_t)}))\in\mathbb{R}^{2J\times 2J}$. Note that the diagonal covariance matrix is due to \textbf{Proposition 1} in section \ref{section:WSS}. In practice, only indices that correspond to the frequency of interest are used. For the tapered case, we assume that the observation $y_t^{(m,F)}$ is generated by the same state-dependent covariance matrix $\Sigma_{s_t}$.
\begin{equation}
p\Big(y_t^{(m,F)}\Big\vert \theta^{(s_t)}\Big)=\mathcal{N}\big(\bar{\text{\bf{0}}},\Sigma_{s_t}\big)\quad\text{ for }\,\,m=1,\cdots,M
\end{equation}
For the prior distribution on $\theta^{(s_t)}$, we use the Inverse Gamma distribution, due to the conjugacy between the normal likelihood and the Inverse Gamma prior.

\subsubsection{HDP-HMM construction}
%a 3-dimensional tensor $y_{t}^{(i,j,k)}$ comprising the real and imaginary components ($j \in \{1,2\}$) of each frequency of interest $i \in \{1,...,f\}$ for each taper $k \in \{2,...,M\}$. 
We model $\{y_t^{(m,F)}(w_j)\}_{t=1}^{T}$ as the realization of a generative process driven by a set of local latent variables $\{s_{t}\}_{t=1}^{T} \in \{1,...,K\}$ that we term "sleep states". Sleep state modeling for each subject proceeds via the following construction for our HDP-HMM:

\begin{align}
\pmb{\beta} \sim \text{GEM}(\gamma), \hspace{0.25cm} \pmb{\pi}_{k}|\beta \sim \text{DP}(\alpha, \pmb{\beta}),\hspace{0.25cm}  b_{j}^{(s_t)} \sim \text{Gamma}(a_{0},b_{0}),\hspace{1cm}\\ 
%\sigma_{k}^{i} \sim \text{IG}(\alpha_{k},\beta_{k}^{i}), 
f(w_{j}^{(s_t)})\sim\text{IG}(a,b_{j}^{(s_t)}),
\hspace{0.25cm} s_{t}|s_{t-1} \sim \text{Multinomial}(\pi_{s_{t-1}}), \hspace{0.25cm} y_{t}^{(m,F)}|s_{t} \sim \mathcal{N}(\bar{\text{\bf{0}}},\Sigma_{s_{t}}) \hspace{0.25cm} 
\end{align}

where IG denotes the Inverse Gamma distribution for each frequency and state specific $f(w_{j}^{(s_t)})$ as outlined in section 2.3.1. To capture the frequency-level characteristics, we place an uninformative Gamma prior over the Inverse Gamma hyperparameter $b_{j}^{(s_{t})}$. Following standard convention, $\pmb{\beta} \sim \text{GEM}(\gamma)$ denotes the stick-breaking construction for the Dirichlet Process (DP) given by 
\begin{align}
\beta_{k} = \beta_{k}'\prod_{i=1}^{k-1}(1-\beta_{i}'),
\end{align}
and $\beta_{k}' \sim \text{Beta}(1,\gamma)$. $\{\beta_{k'}\}_{k'=1}^{K}$ represents a prior over the transition probabilities into a state $k'$, and $\pmb{\pi}_{k} = \{\pi_{k,k'}\}_{k'=1}^{K}$ represents the transition probability from state $k$ to $k'$. Our HDP-HMM is fully nonparametric when we take the limit as $K \rightarrow \infty$ in the GEM stick breaking construction. However, in our work, it is reasonable to truncate $K$ since many poorly represented states will not necessarily confer any additional neurophysiological context, especially considering the finite nature of the data. Finally, we place uninformative gamma hyperpriors over the HDP hyperparameters $\gamma$ and $\alpha$: $\gamma \sim \text{Gamma}(\alpha_{\gamma},\beta_{\gamma})$ and $\alpha \sim \text{Gamma}(\alpha_{\alpha},\beta_{\alpha})$.

\section{Inference}
%Markov chain Monte Carlo (MCMC) approximation methods such as Gibbs sampling are convenient inference tools, since they make asymptotically exact samples from posterior distributions that are otherwize analytically intractable. Unfortunately, Gibbs sampling approaches for HDP-HMMs demonstrate slow convergence, typically due to poor chain-mixing and strong autocorrelations between consecutive time-series observations. 
\subsection{Subject-wise Inference - State Trajectory Sampling}
For model inference, we utilize beam sampling \cite{Gael2008BeamModel}, which integrates slice sampling and dynamic programming to sample whole state trajectories from the model posterior.  HDP-HMMs are infinite-dimensional by formulation, which complicates the computation of state trajectories using ordinary dynamic programming schema. By iteratively sampling an auxiliary variable $u_{t} \sim \text{Uniform}(0,\pi_{s_{t-1},s_{t}})$, the beam sampler imposes a constraint $\pi_{s_{t-1},s_{t}} \geq u_{t}$ on sampled state transitions $p(s_{t}|y_{1:t}^{(m,F)},u_{1:t})$, and consequently sets an upper bound on the set of dynamically computed state trajectories (full details outlined in  \cite{Gael2008BeamModel}). For the posterior distribution of $f\big(w_j^{(s_t)}\big)$, we use following facts: 1) the posterior distribution is also an Inverse Gamma distribution due to conjugacy and 2) there are $M$ pseudo-observations for the DFT coefficients.
\begin{equation}\label{eq:psd_posterior}
p\Big(f\big(w_j^{(s_t)}\big)\Big\vert \{d_{j,k}^{Re,(m)},d_{j,k}^{Im,(m)}\}_{m=1}^M\Big) = \frac{1}{2}\text{IG}\Big(a_j+M,b_{j}^{(s_t)}+\frac{1}{2}\sum_{m=1}^M\Big\{\big(d_{j,k}^{Re,(m)}\big)^2+\big(d_{j,k}^{Re,(m)}\big)^2\Big\}\Big)
\end{equation}
with a gamma prior over the cluster and frequency-specific parameter $b_{j}^{(s_t)}$.  
% Conveniently, the diagonalized covariance matrix effectively reduces overall inference to just $f$ samples from an inverse-gamma posterior.  

\subsection{Group Level Inference - Clustering Analysis}\label{section:clustering}
Following subject-level inference, we perform posthoc clustering analysis to investigate the shared set of states across different subjects. Since the HDP-HMM is modeled separately for each individual, there is no guarantee that any pair of states from different subjects share similar spectral characteristics.

\paragraph{Symmetric KL divergence as the distance metric}
We use the symmetric KL divergence between the Gaussian likelihoods \cite{Shumway2017TimeApplications}, with the choice of the likelihood justified by the asymptotic normality from \textbf{Proposition 1}. Let $c=1,\cdots,C$, denote the cluster index and $f_c$ the spectral characteristic of the cluster $c$. Likewise, let $\widehat{f}_{s_p}$ be the spectral characteristic of the state $s_p$ of the subject $p$. This is chosen to be the MAP estimate of $p\Big(f_{s_p}\Big\vert \theta\Big)$ in eq. (\ref{eq:psd_posterior}). If the observation $y^{(F)}$ is generated from state $s_p$, the log-likelihood is given as (modulo the constants)
\begin{equation}
\log p\Big(y^{(F)}\Big\vert\widehat{f}_{s_p}\Big)=\sum_{w_j}\Big[-\log \widehat{f}_{s_p}(w_j) - \lVert y^{(F)}\rVert^2/\widehat{f}_{s_p}(w_j) \Big]
\end{equation}
If $s_p$ is clustered into cluster $c$, the log-likelihood with respect to $f_c$ would be
\begin{equation}
\log p\Big(y^{(F)}\Big\vert f_c\Big)=\sum_{w_j}\Big[-\log f_c(w_j) - \lVert y^{(F)}\rVert^2/f_c(w_j) \Big]
\end{equation}
Taking the expectation of the log-ratio of the likelihoods, we obtain both sets of KL divergence
% \begin{equation}
% E_{p(y_{s_p}^{(F)}\Vert \widehat{f}_{s_p})}\Big[\lVert y_{s_p}^{(F)}\rVert^2\Big]=\widehat{f}_{s_p}\text{ and }E_{p(y_{s_p}^{(F)}\Vert f_c)}\Big[\lVert y_{s_p}^{(F)}\rVert^2\Big]=f_c
% \end{equation}
\begin{equation}
\begin{split}
I(\widehat{f}_{s_p}; f_c)&=E_{p(y^{(F)}\vert \widehat{f}_{s_p})}\Bigg[\log\frac{p(y^{(F)}\vert \widehat{f}_{s_p})}{p(y^{(F)}\vert f_c)}\Bigg]=\sum_{w_j}\Bigg\{-\log\Bigg(\frac{\widehat{f}_{s_p}(w_j)}{f_c(w_j)}\Bigg)+\frac{\widehat{f}_{s_p}(w_j)}{f_c(w_j)}-1\Bigg\}\\
% I(f_c;\widehat{f}_{s_p})&=E_{p(y^{(F)}\vert f_c)}\Bigg[\log\frac{p(y^{(F)}\vert f_c)}{p(y^{(F)}\vert \widehat{f}_{s_p})}\Bigg] = \sum_{w_j}\Bigg\{-\log\Bigg(\frac{f_c(w_j)}{\widehat{f}_{s_p}(w_j)}\Bigg)+\frac{f_c(w_j)}{\widehat{f}_{s_p}(w_j)}-1\Bigg\}\\
\end{split}
\end{equation}
with $I(f_c;\widehat{f}_{s_p})$ computed similarly. The symmetric KL divergence, $J(\widehat{f}_{s_p};f_c)$, is defined as follows
\begin{equation}\label{eq:clusterdistance}
J(\widehat{f}_{s_p};f_c) = I(\widehat{f}_{s_p}; f_c) + I(f_c;\widehat{f}_{s_p}) = \sum_{w_j} \Bigg\{\frac{\widehat{f}_{s_p}(w_j)}{f_c(w_j)}+\frac{f_c(w_j)}{\widehat{f}_{s_p}(w_j)}-1\Bigg\}
\end{equation}

\paragraph{Weighted K-means clustering}
With eq.(\ref{eq:clusterdistance}) as the distance metric, we now perform K-means clustering on all the states across the subjects, with $f_c$ as the centroid of the cluster $c$. Furthermore, we weight \cite{Tseng2007PenalizedData} $J(\widehat{f}_{s_p};f_c)$ by $n_{s_p}$, the number of occurrences of $s_p$ in subject $p$. $n_{s_p}$ effectively acts as another covariate for clustering the heterogeneous states. For instance, high $n_{s_p}$ would indicate that the state is likely to be one of the canonical sleep stages (REM or NREM). 

We perform two additional preprocessing steps for the clustering analysis. First, we normalize power within each state, $\widehat{f}_{s_p}'(w_j)=\widehat{f}_{s_p}(w_j)/\sum_{w_j}\widehat{f}_{s_p}(w_j)$, to prevent the total power from affecting the clustering. The distance metric in Eq.(\ref{eq:clusterdistance}) has a tendency to assign a high power state to a high power cluster, and vice versa, regardless of the relative power distribution between frequency bands. After normalization, the algorithm is able to cluster the heterogeneous states based on a shared spectral signature. As a second procedure, we take the median value of $n_{s_p}$ across posterior samples as the representative number of occurrences for the specific state for the patient. We observe that after the burn-in period, $n_{s_p}$ is quite consistent, which justifies our choice of the median.

\section{Datasets}
\paragraph{Simulated Sleep-Inspired Data} To test the robustness of our model, we simulate EEG time series with realistic spectral content and temporal dynamics. We do not try to reproduce the extremely complex short-time dynamics of sleep EEG; rather our simulated data gauges the model's capacity to extract a set of discrete spectral characteristics from time-series data. Simulated data is generated by combining the oscillation components time series decomposition method \cite{Matsuda2017MultivariateComponents} and the sleep dynamics \cite{Prerau2017SleepAnalysis}. Our state-space model is described as follow. For $l=1,\cdots,J$

\begin{equation}
x_{l+1}=Rx_{l} + v_l \in \mathbb{R}^{2D} ,  \text{  }
y_{l}=\left(\begin{matrix}1&0&1&...&1&0\end{matrix} \right) x_{l} + w_l \in \mathbb{R}
\end{equation}

where $D$ is the number of frequency components, $v_t \sim \mathcal{N}(\bar{\text{\bf{0}}},Q)$ and $w_t \sim \mathcal{N}(0,\sigma_r^2)$. We define $R$ as a block diagonal matrix composed of 2 by 2 rotation matrices $R_{i}$ where $F_s$ is the sampling frequency, and $f_{i}$ is a frequency (in Hz) of interest in the range $i=1,...,D$ and : 
\begin{equation}
R_i= a_i 
\left(\begin{matrix}
cos(2 \pi f_i / F_s) && sin(2 \pi f_i / F_s) \\
sin(2 \pi f_i / F_s) && -cos(2 \pi f_i / F_s) \\
\end{matrix} \right) \text{ where } 0<a_i<1,
\end{equation}
Our observation PSD is the sum of the observation noise $w_t$ and the PSD of each  oscillation. For any given frequency, the latter peaks around $f_i$ and is fully determined by $a_i, f_i$ and $Q_{2i,2i}$ (the $2i^{th}$ diagonal element of Q) \cite{Matsuda2017MultivariateComponents}. Specific details on frequency bands modeled and the associated PSD for a typical 15s window of simulated data are available in the Supplementary Material Section 2.

%The sleep inspired sequences were generated with D=5. We modeled a slow oscillation ($f_{so} = 0.3$Hz), delta ($f_{\delta} =3$ Hz), theta ($f_{\theta} =5$Hz), alpha ($f_{\alpha} =10$Hz) and sigma ($f_{\sigma} =14$Hz). A typical 15 second generated signal and its associated PSD are reported Figure \ref{fig:simulation} top panels. The other states are illustrated in supplementary figures.

\begin{figure}
  \centering
  \includegraphics[width=\linewidth, trim= 0cm 0cm 0cm 0cm, clip]{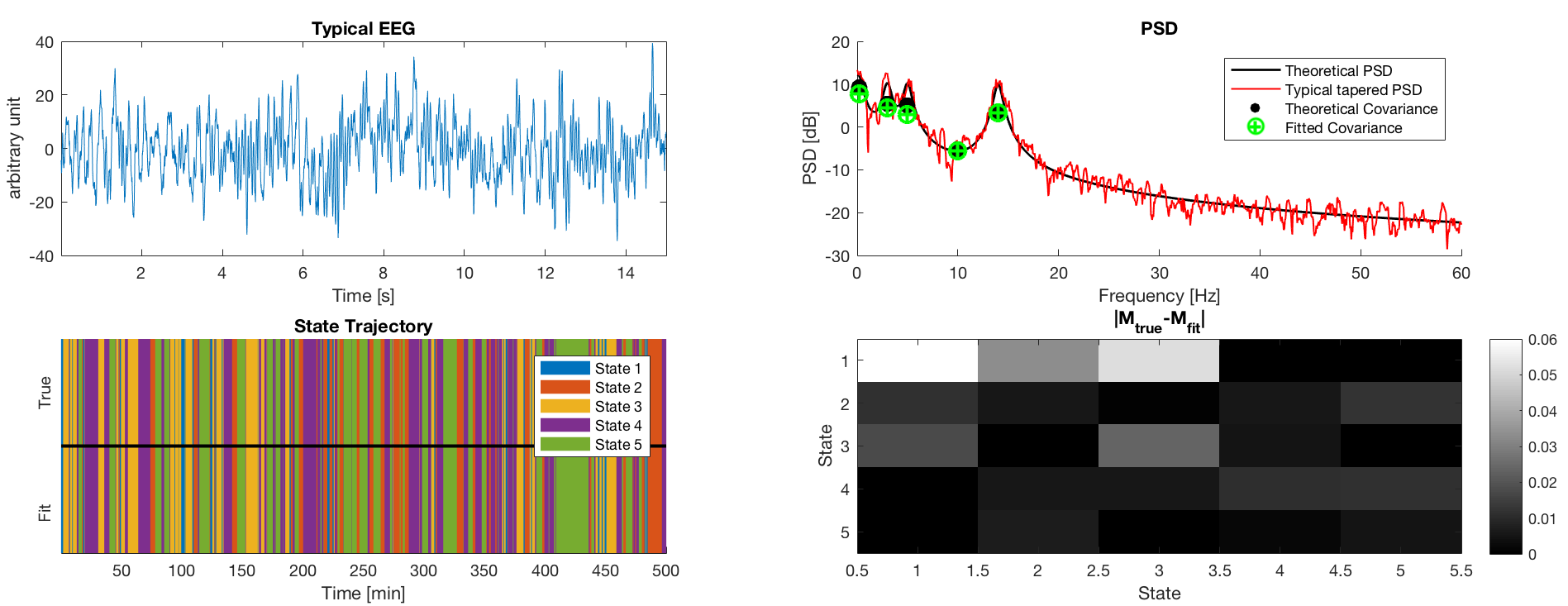}
  \caption{\label{fig:simulation} Simulated Data. For each time window, an EEG signal was generated according to the current hidden state and its corresponding state-space model parameters. A typical EEG trace for one state is depicted on the top left panel. Its multitapered PSD is plotted on the top right panel (orange). The theoretical PSD (black) is determined by the state-space model parameters. The simulated and sampled state trajectories are presented on the bottom left panels. The 5 different states alternate according to a predefined transition matrix, and the difference between simulated and sampled transition matrices are represented on the bottom right panel.}
\end{figure}

\paragraph{Sleep Study Data}
Our experimental data consists of 200Hz high-density (64-channel) EEG recordings from nine healthy right-handed subjects (full clinical details in the original study \cite{Prerau2017SleepAnalysis}). We used the occipital lobe (channel O1) recordings in accordance with a subject-matter expert, who selected the channel to assign canonical sleep stages to 30-second windows. The observations for our model consist of 15-second windows, intended to delineate high frequency sleep phenomena such as sleep ripples and spindles from the signal. Windows containing a total power greater than the 95th percentile of the entire signal were rejected from further analysis. Observation data consisted of multitapered spectral observations (5 tapers, Time Bandwith=4) in $[0.5-2.5]$Hz, $[2.5-4.5]$Hz, $[4.5-6.5]$Hz, $[6.5-8.5]$Hz, $[10.5-12.5]$Hz, and $[12.5-35]$Hz, where the increased resolution at the slow and alpha frequencies is believed to assist the stratification of the NREM1, NREM2, and NREM3 sleep stages \cite{Prerau2017SleepAnalysis}. Specifically, for each tapered window, we take the DFT and examine the power in each frequency band to determine the frequency index corresponding to the median. DFT coefficients corresponding to the frequency index are selected as the spectral observations in the freqency band.

\section{Results}
\subsection{Recovering Sleep State Trajectories}
%A principle line of inquiry in our study focuses on recovering the generative process underlying each observation. We first interrogate model robustness by fitting our HDP-HMM to batches of simulated sleep-like data with fixed ground-truth latent variables. Following this, we use our framework to perform an exploratory analysis of time-series data derived from a well known sleep study.

For each run of the beam sampler, we initialize the following HDP hyperprior distributions: $\gamma \sim \text{Gamma}(1,1)$, $\alpha \sim \text{Gamma}(1,1)$ and $b^{(k)}_{j} \sim \text{IG(1,1)}$. We "burn-in" the first 2000 posterior distribution samples, and draw a subsequent 100 samples for the simulation and 1000 samples for the real data with a step size of 50 to minimize inter-sample autocorrelation.

\paragraph{Simulated Sleep-Inspired Data}

A typical simulation result is presented in Figure \ref{fig:simulation} in which we simulate 2000, 15-second time windows. For each sample from the posterior, our model successfully recovers the correct number of states and the exact state trajectory. Furthermore, the ground truth transition matrix is sampled almost perfectly. Finally, our model recovers the discrete spectral characteristics of the different simulated states. The estimated covariances matrices correspond to the median of the theoretical PSD for each frequency band. As expected, the use of the multitaper framework resulted in a smaller variance in the estimated PSD when compared to a standard DFT approach. More detailed comparison can be found in the Supplementary Material Section 2.

\paragraph{Human Subject EEG}

\begin{figure}
\centering
{\includegraphics[width=\linewidth]{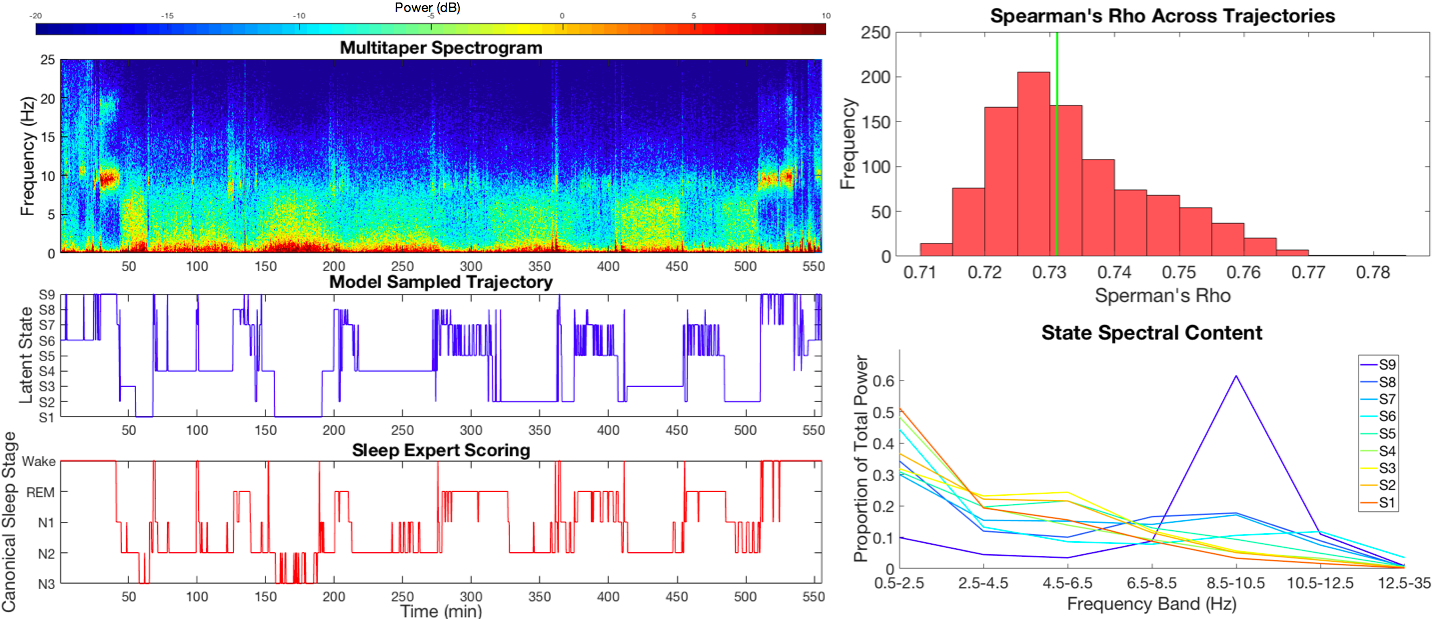}}
\caption{\label{fig:stateTrajectories}Top left: A multitaper spectogram across the entire sleep study duration for Subject 5. Middle left: A state trajectory chosen from the posterior distribution samples of the HDP-HMM according to the median Spearman rho estimate. Bottom left: Sleep expert hypnogram. Top right: Histogram of Spearman's rho values computed between all sampled state trajectories and the hypnogram, where the green vertical line highlights the median of the distribution. Bottom right: The estimated power spectral density for each state in the state trajectory.}
\end{figure}

Hypnograms illustrate canonical sleep stage traces in terms of descending levels of arousal, where a 5 on the hypnogram corresponds to the highest level of arousal (wake), and a 1 to the lowest (NREM3). For comparison, our state labels are also reordered, using descending state-wise alpha band power as a proxy for arousal. Spearman's rho $\rho_{t}$ is used to compute the similarity in temporal dynamics between sampled state trajectories and the hypnogram. We present Subject 5 as a case study in Figure 3, where all sampled state trajectories are used to construct a distribution over $\rho_{t}$ values. The distribution is tightly localized around a median value of $\rho_{t} = 0.73$, demonstrating that the model effectively captures the transition dynamics in the hypnogram. Remarkably, transitions between NREM and REM states in the hypnogram are mirrored by similar transitions between slow states S1, S2 and S3 to REM-like states such as S5, S7 and S8 in the state trajectory. Similarly, inter-NREM transitions in the hypnogram coincide with synonymous transition events in the state trajectory between S1, S2 and S3. Our model introduces additional "micro"-states beyond the standard canonical sleep stages, typically coinciding with more volatile dynamics in the spectrogram. This can clearly be discerned in Figure 3, where epochs of REM in the hypnogram are broken into alternating S8/S7 (characteristic REM PSDs) and S5 (slow dynamics dominated PSD) in the state trajectory. This is unsurprising considering that a) sleep consists of highly heterogeneous and fluctuating oscillatory dynamics, b) current defined canonical sleep stages are unable to account for inter-subject variability and c) both additional and combined sleep stages have been reported \cite{Ulrich2014AnalysisTime-Series,Prerau2017SleepAnalysis}. 
When considering the variation introduced from additional subject-specific micro-states, the general temporal dynamics are recovered well across all the subjects with an average distribution median value of $\rho_{t} =  0.69$.

\subsection{Clustering Sleep States Across Subjects}

\begin{figure}
\centering
\includegraphics[width=\linewidth, trim= 0cm 0cm 0cm 0cm, clip]{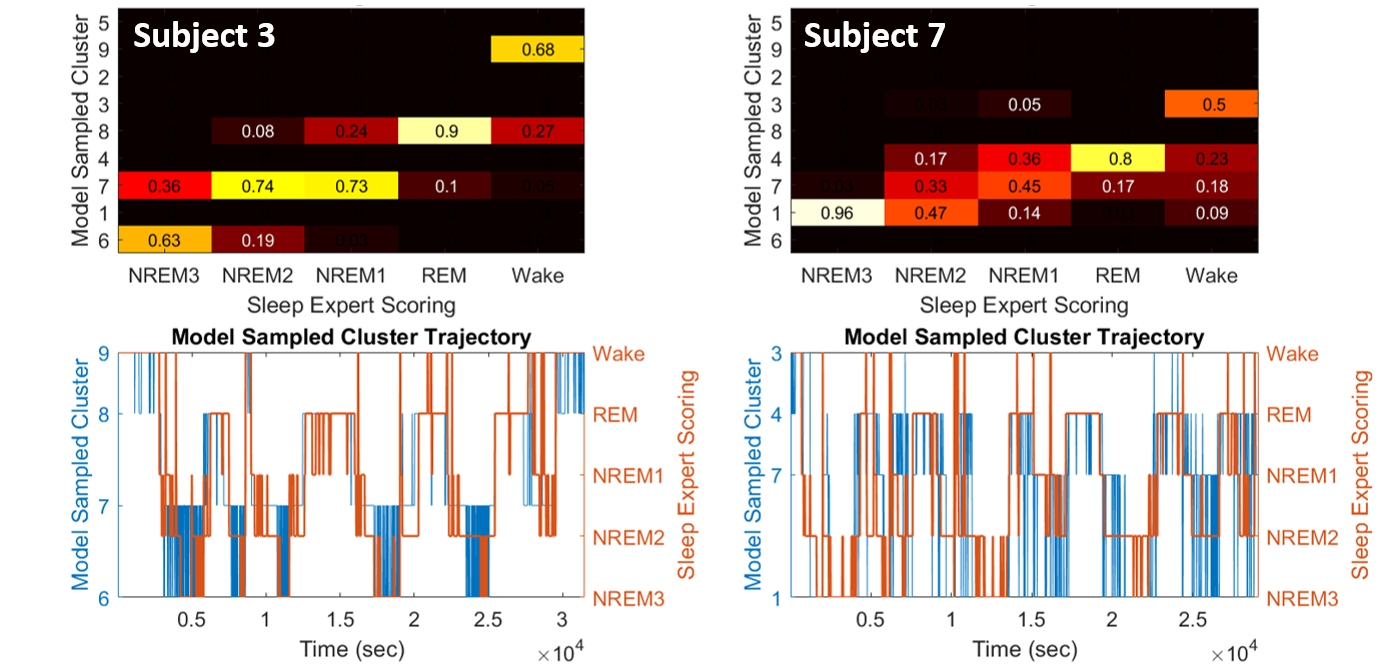}
\caption{\label{fig:clusterAnalysis} Detailed Cluster Analysis for Two Subjects. Top: Heatmaps illustrating the proportion of time spent in each sleep expert scored stage belongs to each cluster. Bottom: The cluster trajectories on the right show the temporal dynamics of the clusters compared to the sleep expert scored stages. }
\end{figure}

%The clustering analysis aims to group states based on the underlying hypothesis that every subject has both "REM-like" and "NREM-like" states; however, their specific spectral characteristics may vary. Across our nine subjects, a total of 103 states and nine clusters were recovered. A discussion of the cluster spectral characteristics and comparison with sleep expert scored stages can be found in the Supplementary Material. 

The clustering analysis aims to group states based on the underlying hypothesis that every subject has both "REM-like" and "NREM-like" states; however, their specific spectral characteristics may vary. Across the nine subjects, a total of 103 states and nine clusters were recovered. The HDP-HMM recovered sleep states for each subject are then mapped to their respective clusters. Finally, the cluster trajectory and hypnogram are compared for each subject. We discuss their spectral characteristics in the Supplementary Material Section 3.1. 

Although nine clusters were identified, not all clusters were visited by each subject. An example of the heterogeneity in cluster dynamics across subjects is shown for two subjects in Figure \ref{fig:clusterAnalysis}. The heatmaps link the proportion of time spent in each sleep expert scored stage to the time spent in each cluster. The clusters are organized from bottom to top in order of increasing power in the alpha band (8-12 Hz). For example, for Subject 3, the fourth column (REM) indicates that 90 percent of the windows labeled REM by the sleep expert were classified as belonging to Cluster 8 and the remaining 10 percent as belonging to Cluster 7. The cluster trajectory plots in Figure \ref{fig:clusterAnalysis} show the temporal dynamics of the clusters compared to the sleep expert scored stages.

There are several noteworthy observations. Firstly, across all subjects, including the two in Figure \ref{fig:clusterAnalysis}, there is a clear shift towards increasing power in the alpha band from deep sleep (NREM3) to lighter sleep (NREM1), REM, and then wake. This is evidenced by increasing proportions of time spent in higher clusters going from left to right on the heatmaps, which agrees with known sleep stage physiology (\cite{Prerau2017SleepAnalysis}). Secondly, certain clusters are dominant during the same scored sleep stage across subjects, while others modulate their behavior between subjects. For example, across all nine subjects, Clusters 1 and 6 are only dominant during the deepest sleep states (NREM3 and NREM2), while  Clusters 3,5 and 9 are only dominant during wake states. However, Cluster 4 is dominant during NREM2, NREM1, and REM in different subjects, which suggests that each person may have a slightly different spectral signature for the same scored sleep stage and that certain combinations of network activity may not be relegated solely to one stage. (For more details on cluster dominance during scored sleep stages, see Table S1 in the Supplementary Material Section 3.2.)

Finally, both the heatmaps and cluster trajectories demonstrate that there are faster sub-oscillations occurring within each sleep expert scored stage across subjects. However, these sub-oscillations occur at different times across subjects. For example, Subject 3 oscillates rapidly between clusters during deep sleep (NREM3), but is very stable during REM. On the other hand, Subject 7 is stable during deep sleep but oscillates more during lighter stages of sleep and wake. This stage-dependent oscillatory heterogeneity can been seen clearly in Table S2 (Supplementary Material Section 3.2), which has the computed transition rates per minute for each sleep stage for both subjects. In a broader context, stage-dependent sub-oscillations may help describe heterogeneity across subjects and populations.

\section{Conclusion}
Overall, this method provides a robust, Bayesian nonparametric framework for identifying the spectral content, number, and dynamics of salient recurring oscillatory states during sleep across patients. In addition, we identify subject-specific "microstates" within canonical sleep stages and furthermore, discover stage-dependent sub-oscillations with shared spectral signatures across subjects. This work can serve as a basis for novel mechanistic studies focusing on the network dynamics of these states, as well as their clinical and scientific relevance. By liberating sleep from the practical necessities of human scoring and a priori assumptions of machine learning algorithms, we pave the way for a higher dimensional, data-driven feature space for biomarker detection and clinical intervention.

\newpage 

\newpage
\small
\bibliography{mendeley}
\bibliographystyle{unsrt}

\end{document}